  \documentclass[letterpaper, 10 pt, conference]{ieeeconf}

  \IEEEoverridecommandlockouts
  \usepackage{amssymb}
  \usepackage[dvipsnames]{xcolor}
  \usepackage{color, colortbl}
  \usepackage{booktabs}
  \usepackage{graphicx}
  \usepackage{amsmath}
  \usepackage{multirow}
  \usepackage{caption}
  \usepackage{dsfont}
  \usepackage{cite}
  \usepackage{algorithmic}
  \usepackage{algorithm}
  \usepackage{textcomp}
  \usepackage{hyperref}
  \hypersetup{
    colorlinks=true,
    linkcolor=black,
    citecolor=black,
    urlcolor=NavyBlue,
    pdfborder={0 0 0},
  }
  \newcommand{\psmproject}{https://simeon-ned.github.io/predictive-style-matching/}

  \title{\LARGE \bf
  Predictive Style Matching:\\
  Natural and Robust Humanoid Locomotion
  }

  \author{Simeon Nedelchev$^{1,2}$, Ekaterina Chaikovskaia$^{1}$, Egor Davydenko$^{1}$, Eduard Zaliaev$^{3}$, and Roman Gorbachev$^{1}$
  \thanks{This research was supported by the Ministry of Economic Development of the Russian Federation (Agreement No. 139-15-2025-013 with MIPT, dated June 20, 2025. IGK 000000C313925P4B0002).}
  \thanks{$^{1}$Moscow Institute of Physics and Technology (MIPT), Dolgoprudny, Moscow Region, Russia}
  \thanks{$^{2}$Innopolis University, Innopolis, Russia}
  \thanks{$^{3}$Sber Robotics Center, Moscow, Russia}
  \thanks{Project page, videos, and supplementary material: \url{\psmproject}.}
  }

\begin{document}

  \maketitle

  %%%%%%%%%%%%%%%%%%%%%%%%%%%%%%%%%%%%%%%%%%%%%%%%%%%%%%%%%%%%%%%%%%%%%%%%%%%%%%%%
  \begin{abstract}
  Reinforcement learning has become the prevailing approach to humanoid locomotion control: policies transfer reliably from simulation to hardware and recover gracefully from disturbances.
  Motion quality, however, still lags behind: task-only rewards often converge to stiff, asymmetric gaits, while motion imitation methods improve appearance but become more sensitive to external disturbances because reference signals can oppose the transient poses needed to regain balance.
  We propose \href{\psmproject}{\emph{Predictive Style Matching}} (PSM), in which an offline predictor maps the robot's lower-body state history and velocity commands to interpretable upper-body joint and gait targets that shape the rewards during training.
  Because the targets are state-conditioned rather than time-indexed and the predictor is used only at training time, the deployed controller inherits the proprioceptive interface and inference cost of a task-only RL baseline.
  On the Unitree G1, in both simulation and hardware, PSM reduces upper-body style error by roughly an order of magnitude over task-only RL while preserving its fall-recovery rate, whereas the motion-imitation baseline attains the lowest style error but fails to recover from disturbances about five times as often.
  \end{abstract}

  %%%%%%%%%%%%%%%%%%%%%%%%%%%%%%%%%%%%%%%%%%%%%%%%%%%%%%%%%%%%%%%%%%%%%%%%%%%%%%%%
  \section{Introduction}

  Reinforcement learning (RL) has become a standard tool for contact-rich legged locomotion.
  Massively parallel simulation now trains policies in minutes and seamlessly transfers them to hardware~\cite{rudin2022learning,zakka2026mjlab}. Recent RL-based controllers can recover from pushes, traverse difficult terrain, and track increasingly dynamic whole-body behaviors~\cite{lee2020learning,miki2022learning,cheng2024extreme,hoeller2024anymal,rudin2024parkour,he2024asap,shi2024kungfubot,sleiman2026zest,ha2023learning}.
  Yet hardware capability has outpaced motion quality: velocity and balance tracking rewards rarely specify how the torso, arms, and pelvis should coordinate during ordinary walking.

  If left underspecified, task-only policies converge to stiff, asymmetric gaits that maximize reward but look unnatural~\cite{escontrela2022adversarial,rudin2022learning}.
  Hand-tuned shaping penalties and reward curricula~\cite{rudin2022learning,ji2022concurrent} mitigate the symptoms but do not encode a compact model of whole-body human coordination.

  \begin{figure}[t]
    \centering
    \href{https://www.youtube.com/watch?v=9VCARWkgY-s&feature=youtu.be}{%
      \includegraphics[clip, trim=60 60 80 80, width=\columnwidth]{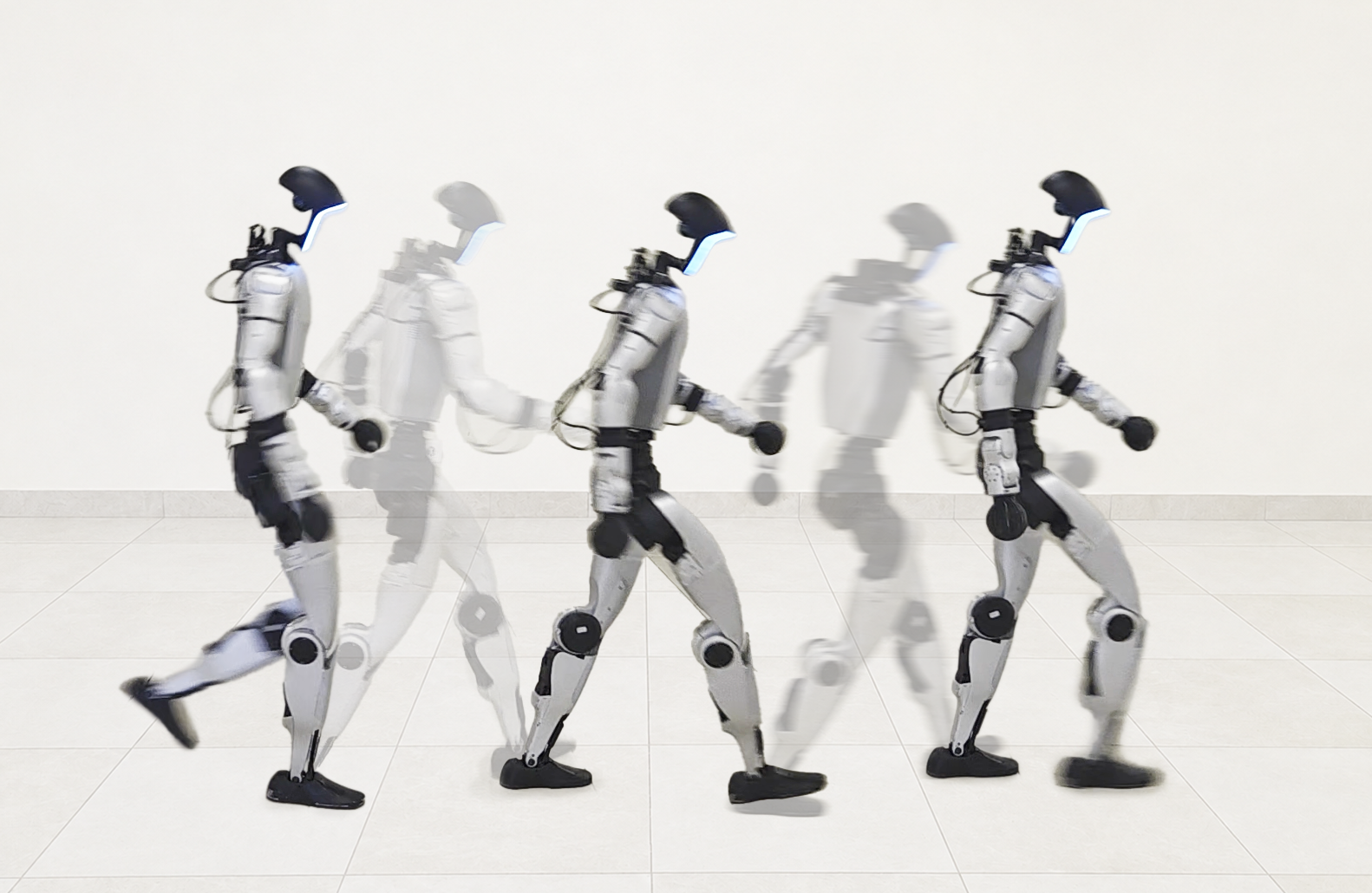}%
    }
    \caption{Human-like whole-body locomotion on the Unitree G1 with PSM (hardware).
    Under a fixed velocity command, the deployed policy autonomously produces coordinated arm swing, pelvis motion, and natural stepping, while retaining the inference cost of a vanilla task-only controller}
    \label{fig:psm-hardware}
  \end{figure}

  Human motion data is a more principled source of regularization, but its use during training and deployment matters.
  Motion imitation methods such as DeepMimic~\cite{peng2018deepmimic} and its successors (e.g., ZEST, ASAP)~\cite{sleiman2026zest,he2024asap,gu2024videomimic} drive RL training with time-indexed pose matching against motion-capture or video clips; deployed policies use phase signals in the observation space or explicit reference inputs to continue tracking the task rewards when balance is eventually regained.
  Adversarial motion priors (AMP)~\cite{peng2021amp,escontrela2022adversarial} avoid a reference clock by scoring simulated state transitions with a learned discriminator, but the feedback is a scalar over the full state: it is not tied to specific joints or interpretable gait statistics, and it constrains contacts and leg motions to strictly follow task rewards during recovery.

  \begin{figure*}[!t]
    \centering
    \includegraphics[width=0.9\textwidth]{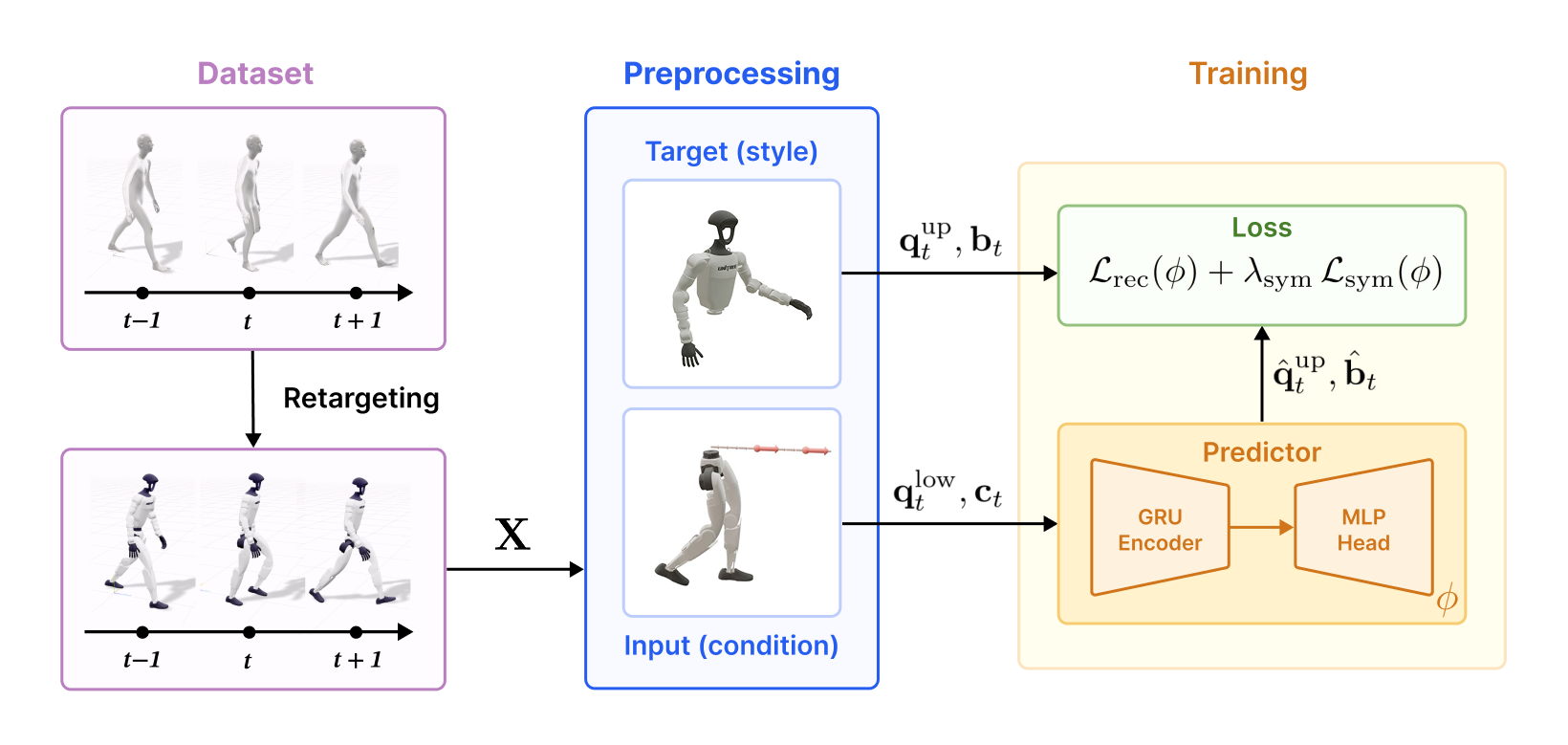}
    \caption{Stage 1 (predictor training): offline supervised learning of $f_\phi$ on retargeted dataset clips. Lower-body history and command cues are preprocessed into conditioning inputs, while upper-body joints and gait descriptors provide targets; the GRU--MLP predictor is trained with reconstruction and left--right symmetry losses, then frozen.}
    \label{fig:stage1-predictor}
  \end{figure*}

  A complementary line of work from physics-based character animation and game development learns \emph{generative} motion models---diffusion, autoregressive, and adversarial latent policies---for rich kinematic synthesis and interactive steering~\cite{shi2024amdm,tevet2024condmdi,tevet2025closd,tessler2023calm,peng2022ase}.
  Coupled with simulation, these models increasingly close the loop to deployable whole-body skills~\cite{tessler2024maskedmimic,liao2025beyondmimic}, but they target a different operating point: generated samples are less direct to convert into named per-joint/gait reward terms, iterative sampling is expensive inside massively parallel RL, and hardware controllers still favor low-latency proprioceptive inference.
  We therefore use a lightweight \emph{discriminative} predictor as a train-time style oracle; unlike imitation, AMP, or generative priors alone, it provides a \emph{command-conditioned map from current lower-body motion} to the upper-body and gait targets needed by a recovery-aware controller.

  This leaves a familiar training-time trade-off: vanilla velocity tracking RL policies are robust under disturbance but have no explicit reward for natural arm swinging or stepping rhythm, while motion imitation prioritizes the desired appearance at the cost of recovery quality~\cite{peng2018deepmimic,sleiman2026zest}.
  The question we address is how to inject human coordination into RL training without adding a reference generator at deployment or forcing the policy to chase style targets that oppose recovery.

  We propose \emph{Predictive Style Matching} (PSM), a two-stage pipeline that keeps rich style supervision in training while preserving a policy-only deployment interface.
  Stage 1 trains an offline predictor $f_\phi$ to map lower-body history and command cues to interpretable upper-body and gait targets; Stage 2 uses frozen predictions as reward terms during PPO training, so deployment remains the same proprioceptive actor as vanilla RL, without running $f_\phi$.
  Concretely, this paper contributes:
  \begin{itemize}
    \item \textbf{Motion predictor} $f_\phi$ conditioned on lower-body state and velocity command, trained offline on retargeted human walking data, that emits \emph{interpretable} style targets: 17 upper-body joint angles and a set of gait/body descriptors rather than a time-indexed pose or an abstract scalar discriminator score.
    \item \textbf{Training pipeline} in which the frozen $f_\phi$ outputs are used solely through matching rewards on top of a standard locomotion task reward setup, leaving the deployed controller with the same proprioceptive interface and inference cost as a task-only RL baseline.
    \item \textbf{Controlled comparison.} A simulation and hardware evaluation on the Unitree G1 against vanilla locomotion and a motion-imitation baseline under matched MDP, domain randomization, curriculum, and disturbance schedules, showing that PSM can bridge the naturalness gap between motion-conditioned and robust vanilla policies.
  \end{itemize}

  %%%%%%%%%%%%%%%%%%%%%%%%%%%%%%%%%%%%%%%%%%%%%%%%%%%%%%%%%%%%%%%%%%%%%%%%%%%%%%%%
  \section{Predictive Style Matching}
  \label{sec:method}

\begin{figure*}[!t]
    \centering
    \includegraphics[width=1.05\textwidth]{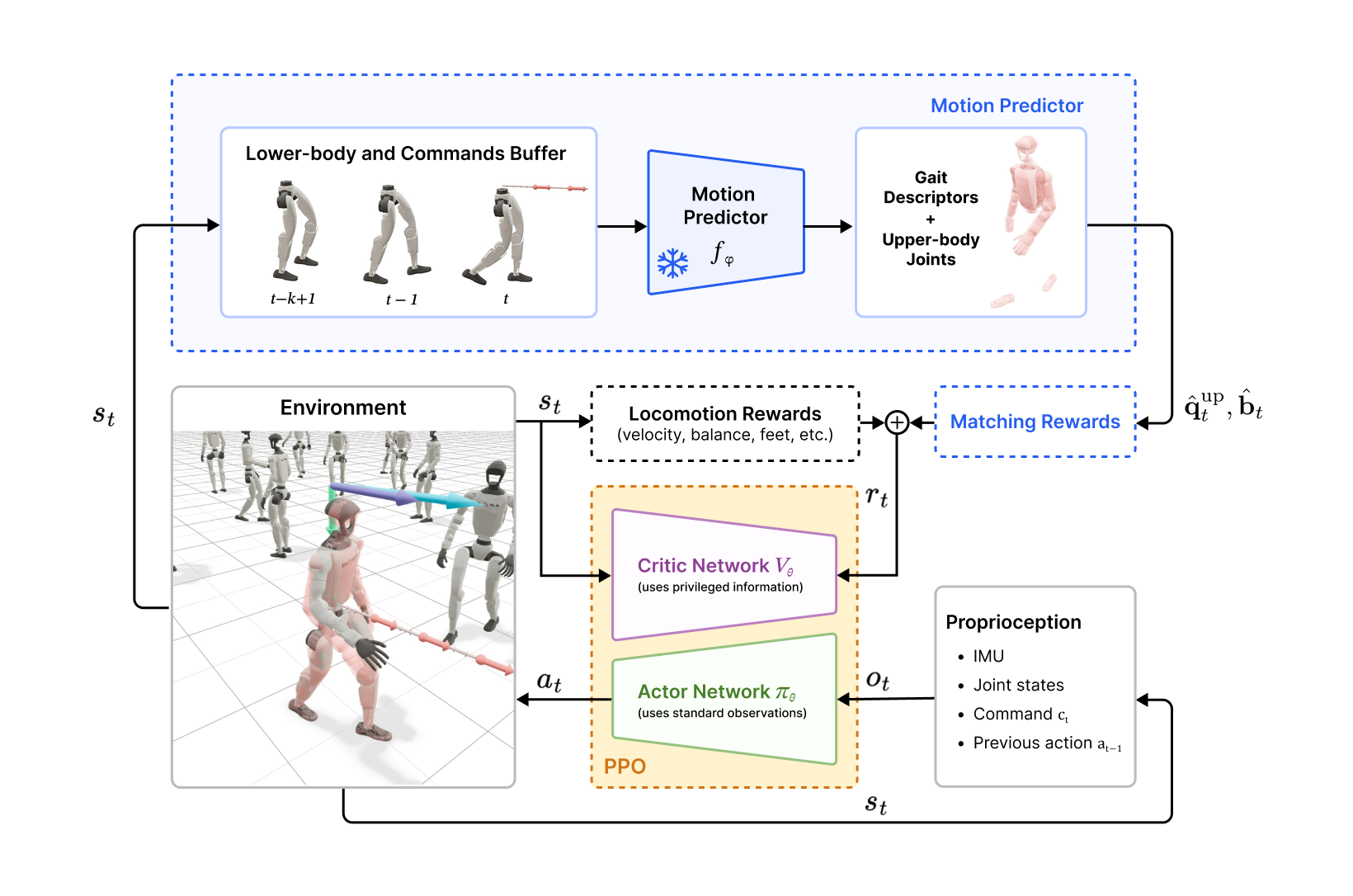}
    \caption{Stage 2 (RL with predictive style matching): PPO trains actor--critic policies in simulation while querying the frozen $f_\phi$ on rolling lower-body and command buffers. Predicted upper-body and gait targets define matching rewards that are added to locomotion rewards during training only; deployment keeps a standard proprioceptive actor without running the predictor.}
    \label{fig:stage2-rl}
  \end{figure*}

  We now turn that motivation into a concrete training pipeline.
  PSM keeps style supervision in training while preserving a policy-only deployment interface in two stages: Stage 1 learns an offline predictor $f_\phi$ from retargeted walking data, and Stage 2 uses its frozen outputs as matching rewards during PPO training of $\pi_\theta$.
  As a result, the deployed controller remains the same proprioceptive actor as vanilla RL, without a phase clock, reference clip, or predictor network at run time.
  This separation lets us compare style quality and disturbance recovery under a matched training setup, rather than changing the deployment interface across methods.
  Figs.~\ref{fig:stage1-predictor} and~\ref{fig:stage2-rl} summarize these two stages, detailed in Secs.~\ref{sec:motion-predictor} and~\ref{sec:rl}.

  \subsection{Motion predictor}
  \label{sec:motion-predictor}

  We begin with the offline stage: $f_\phi$ is fit by supervised learning on retargeted human walking, with no RL in the loop.
  We assemble corpus $\mathcal{D}$ from BoneSeed~\cite{bonesstudio2026aidatasets}: a walking subset ($\sim$10\% of available clips) is retargeted to the Unitree G1 and used to learn the map
  \begin{equation}
    (\hat{\mathbf{q}}^{\mathrm{up}}_t, \hat{\mathbf{b}}_t) = f_\phi\big(\mathcal{H}_t, \mathbf{c}_t\big),
    \label{eq:predictor}
  \end{equation}
  where $\mathcal{H}_t$ is a rolling, normalized history of lower-body joint angles, optional root-frame foot positions, and yaw-aligned root velocities, and $\mathbf{c}_t$ collects short-horizon velocity command cues.
  On the G1, $\hat{\mathbf{q}}^{\mathrm{up}}_t$ contains $N_{\mathrm{up}}{=}17$ joint angles (waist yaw/roll/pitch; left and right shoulder pitch/roll/yaw; left and right elbow; left and right wrist roll/pitch/yaw), and $\hat{\mathbf{b}}_t$ collects step length, step width, left/right foot pitch, left/right foot yaw relative to the root, root height, and predicted duty cycle (double-support fraction).
  These are physically interpretable targets, not a latent style code; a qualitative \href{https://youtube.com/shorts/6y1swXFSS7I?feature=share}{predictor visualization} shows $f_\phi$ forecasting the upper body (red ghost) from lower-body motion alone.

  Inputs and network architecture are specified next (Sec.~\ref{sec:rl} uses $T_h{=}25$, $T_p{=}10$ at 50\,Hz when $f_\phi$ is queried online).
  The rolling history $\mathcal{H}_t$ bundles three synchronized signals over the last $T_h$ control steps: lower-body joint angles $\mathbf{q}^{\mathrm{low}}_t \in \mathbb{R}^{N_{\mathrm{low}}}$ ($N_{\mathrm{low}}{=}8$ hip/knee joints), left and right foot positions in the root frame, $\mathbf{p}^{\mathrm{foot}}_t \in \mathbb{R}^{6}$, and root linear and angular velocity in the yaw-aligned frame, $\mathbf{v}_t{=}(v_x,v_y,\omega_z)$, stacked as $\mathbf{V}_{t-T_h+1:t}\in\mathbb{R}^{T_h\times 3}$.
  The command block $\mathbf{c}_t$ concatenates the current velocity target $(v_x, v_y, \omega_z)$ with short-horizon trajectory cues: for each lookahead $\Delta\in\{15,30,45\}$ frames ($\approx$0.3--0.9\,s), the root-frame displacement, heading direction, and yaw change from $t$ to $t{+}\Delta$.
  A single-layer GRU encodes the per-step sequence $[\mathbf{q}^{\mathrm{low}}, \mathbf{p}^{\mathrm{foot}}]\in\mathbb{R}^{T_h\times(N_{\mathrm{low}}+6)}$ into a 128-D embedding; root velocity is handled separately, with the temporal mean and most recent value of $\mathbf{V}_{t-T_h+1:t}$ forming a 6-D summary that is concatenated with the GRU embedding and $\mathbf{c}_t$.
  A two-layer MLP then regresses $(N_{\mathrm{up}}{+}N_{\mathrm{b}})\cdot T_p$ denormalized targets---$N_{\mathrm{up}}{=}17$ upper-body joints and $N_{\mathrm{b}}{=}8$ gait descriptors at each of the next $T_p$ steps.

  Supervised training minimizes the horizon-averaged squared error on upper-body joints and gait descriptors, regularized by a left--right symmetry term:
  \begin{equation}
  \begin{aligned}
    \mathcal{L}(\phi) = \;& \mathbb{E}_{\mathcal{D}} \frac{1}{T_p}\!\sum_{k=0}^{T_p-1}\!\Big[ \|\hat{\mathbf{q}}^{\mathrm{up}}_{k} - \mathbf{q}^{\mathrm{up}\star}_{k}\|_2^2 \\[-1pt]
        & \qquad \qquad \qquad + \lambda_b\, \|\hat{\mathbf{b}}_{k} - \mathbf{b}^{\star}_{k}\|_2^2 \Big] + \lambda_{\mathrm{sym}}\, \mathcal{L}_{\mathrm{sym}}(\phi),
  \end{aligned}
  \label{eq:predictor-loss}
  \end{equation}
  where $(\mathbf{q}^{\mathrm{up}\star}_k, \mathbf{b}^{\star}_k)$ are the ground-truth targets at horizon step $k$ drawn from $\mathcal{D}$, and $\mathcal{L}_{\mathrm{sym}}(\phi) = \big\| f_\phi(\sigma(\mathcal{H}_t, \mathbf{c}_t)) - \sigma\big(f_\phi(\mathcal{H}_t, \mathbf{c}_t)\big) \big\|_2^2$ enforces equivariance to the left--right mirroring operator $\sigma$.
  Concretely, $\sigma$ reflects the robot across its sagittal plane: left and right joint channels are permuted, with each channel multiplied by $\pm1$ according to its rotation axis (roll/yaw axes flip sign; pitch axes do not), lateral root velocity $v_y$ and yaw rate $\omega_z$ are negated, and trajectory cues that encode lateral displacement or heading change receive the same sign flips.
  We use $\lambda_b{=}1$ and ramp $\lambda_{\mathrm{sym}}$ linearly from $0$ to $0.1$ over the first 2500 updates; input noise (std.\ 0.01 in normalized coordinates) improves robustness to small sim--mocap mismatch, and light feature regularization discourages unrealistic waist motion, foot orientation, and spurious cadence.
  We optimize~\eqref{eq:predictor-loss} with Adam (learning rate $10^{-3}$, batch size $512$, weight decay $10^{-5}$) for 45k updates with gradient-norm clipping at $1.0$.
  On a single RTX~4090, training the predictor takes about 60 minutes on the current subset and scales from minutes to a few hours across the dataset sizes we use; the predictor is then frozen for the rest of the pipeline.

  \subsection{Embedding in RL}
  \label{sec:rl}

  With $f_\phi$ fixed, we now turn to how it shapes policy learning in Stage 2 (Fig.~\ref{fig:stage2-rl}).
  We formulate locomotion as an MDP $(\mathcal{S}, \mathcal{A}, P, r, \gamma)$ and train with on-policy PPO~\cite{schulman2017proximal} in massively parallel MuJoCo simulation~\cite{zakka2026mjlab}.
  The G1 policy maps $\mathbf{o}_t$---IMU, joint state, contacts, command, and the previous action---to joint-position targets executed by low-level PD controllers.
  The shared locomotion reward $r^{\mathrm{loco}}_t$ tracks a resampled twist command $\mathbf{u}_t$ in the root-yaw frame and penalizes orientation error, height drift, foot slip, joint-limit violations, action rate, and impacts.
  Domain randomization covers friction, actuator bias, center-of-mass offset, and external pushes~\cite{rudin2022learning,escontrela2022adversarial}.

  At each simulation step, the lower-body/command buffer is updated, the frozen predictor is queried in receding-horizon mode, and only the first forecast ($k{=}0$) is used as the instantaneous target $(\hat{\mathbf{q}}^{\mathrm{up}}_t, \hat{\mathbf{b}}_t)$ from~\eqref{eq:predictor}; style matching remains disabled until the history buffer has accumulated $T_h$ steps.
  On top of $r^{\mathrm{loco}}_t$, PSM adds per-quantity matching rewards driven by these frozen-predictor outputs.
  For each upper-body joint $j$ and each gait/body descriptor $b_i$ we apply an exponential kernel:
  \begin{align}
    r^{\mathrm{up}}_t       &= \exp\!\left(-\tfrac{1}{N_{\mathrm{up}}}\sum_{j} \tfrac{(q^{\mathrm{up}}_{j,t} - \hat{q}^{\mathrm{up}}_{j,t})^2}{\sigma_j^2}\right),
    \label{eq:r-up} \\
    r^{\mathrm{gait},i}_t  &= \exp\!\left(-\tfrac{(b_{i,t} - \hat{b}_{i,t})^2}{\sigma_i^2}\right),
    \label{eq:r-gait}
  \end{align}
  where the $\sigma$'s set per-quantity tolerances and $b_i$ ranges over step width, step length, foot pitch/yaw, root height, and predicted duty cycle.
  The matching reward and total RL reward are
  \begin{equation}
    r^{\mathrm{match}}_t = w_{\mathrm{up}}\,r^{\mathrm{up}}_t + \sum_i w_i\,r^{\mathrm{gait},i}_t,
    \label{eq:r-match}
  \end{equation}
  \begin{equation}
    r_t = r^{\mathrm{loco}}_t + \lambda_{\mathrm{match}}(k)\,r^{\mathrm{match}}_t(\hat{\mathbf{q}}^{\mathrm{up}}_t, \hat{\mathbf{b}}_t).
    \label{eq:psm-total-reward}
  \end{equation}
  The actor reads only standard proprioception; matching enters exclusively through $r_t$.
  We hold $\lambda_{\mathrm{match}}(k){=}0$ during an initial locomotion-only phase and ramp it up only after balance and recovery behavior are established, so style matching never outweighs the task reward at initialization.
  During the subsequent style curriculum, $w_{\mathrm{up}}$ is raised before the individual gait weights, and the generic hand-designed gait-phase shaping is correspondingly attenuated.
  The policies evaluated in Sec.~\ref{sec:evaluation} use both upper-body and gait terms, including the predicted duty cycle, because explicit gait-descriptor matching proved more stable than relying on upper-body coupling or a fixed phase clock alone.
  Training uses $T_h{=}25$ and $T_p{=}10$ at 50\,Hz control (200\,Hz simulation, decimation 4); only $\pi_\theta$ is exported for deployment, and evaluating $f_\phi$ on the rolling buffers adds about 10\% to RL iteration time on an RTX~4090, while the deployed compute path is identical to vanilla RL.

  Finally, we contrast this construction with two natural alternatives targeting the same problem.
  Clip-tracking baselines replace $\hat{q}^{\mathrm{up}}_{j,t}$ in~\eqref{eq:r-up} with a time-indexed reference pose $q^{\mathrm{ref}}_{j,t}$ read from a clip phase index, so the target is independent of the robot's recovery state; AMP collapses style supervision to a single discriminator score over whole-state transitions, which neither restricts the signal to an upper-body and gait subset nor decomposes it along the per-quantity axes used at evaluation time.
  In this sense, PSM treats style as a state-conditioned prior over feasible recovery behavior, not as a trajectory that must be tracked frame by frame.
  The next section measures both contrasts against PSM under matched training conditions.

  %%%%%%%%%%%%%%%%%%%%%%%%%%%%%%%%%%%%%%%%%%%%%%%%%%%%%%%%%%%%%%%%%%%%%%%%%%%%%%%%
  \section{Evaluation}
  \label{sec:evaluation}

  We evaluate whether PSM improves whole-body naturalness over task-only RL without sacrificing robustness.
  The three policies---vanilla RL, Tracking, and PSM---are tested on Unitree G1 in simulation and on hardware under matched MDP, observation space, PPO budget, randomization, curriculum, and disturbance schedules, so differences are attributable to the style signal.
  Among the three, PSM and vanilla RL share a policy-only proprioceptive deployment interface, whereas Tracking requires a reference phase and a reference pose at run time.
  Because exhaustive trajectory sweeps are impractical on hardware, scenario plots come from a large simulation batch and the robot executes a representative subset of the same $\mathbf{u}(t)$ schedules (Fig.~\ref{fig:psm-hardware} and the supplementary disturbance video).

  All three policies are trained in the same MJLAB stack~\cite{zakka2026mjlab}, with identical PPO settings, observation space, and reward scaffolding, so that only the style signal varies across methods.
  \emph{Vanilla RL} trains with $r_t = r^{\mathrm{loco}}_t$ only; style-matching and reference-tracking weights are zero.
  \emph{Tracking} stands in for the broad family of clip-tracking RL controllers~\cite{peng2018deepmimic,he2024asap,gu2024videomimic}, in which the reward is a time-indexed match against a recorded human pose; concretely, we use the BeyondMimic~\cite{liao2025beyondmimic} tracking implementation that ships with MJLAB~\cite{zakka2026mjlab}, fed the same $\mathcal{D}$ clips as the predictor, and train one tracking policy per reference trajectory.
  At each step the environment advances a clip phase index and exposes a reference pose $q^{\mathrm{ref}}_t$ for the upper body; the imitation reward $r^{\mathrm{imit}}_t$ uses the kernel of~\eqref{eq:r-up} with $\hat{q}^{\mathrm{up}}_{j,t}$ replaced by $q^{\mathrm{ref}}_{j,t}$, and the total reward is $r_t = r^{\mathrm{loco}}_t + \lambda_{\mathrm{imit}}\, r^{\mathrm{imit}}_t$ with the same locomotion-then-style curriculum used for PSM.
  Universal trackers such as ZEST~\cite{sleiman2026zest} can in principle follow many references, but rely on a different observation interface and substantially more training data; we treat them as out of scope for this controlled comparison, whose purpose is to isolate the robustness cost of explicit reference tracking that is already present in single-clip trackers.
  \emph{PSM} uses $r_t = r^{\mathrm{loco}}_t + \lambda_{\mathrm{match}}\, r^{\mathrm{match}}_t$ with targets from the frozen $f_\phi$ during training (Sec.~\ref{sec:rl}) and deploys with the same interface as vanilla RL.

  \begin{figure}[!t]
    \centering
    \includegraphics[width=\columnwidth]{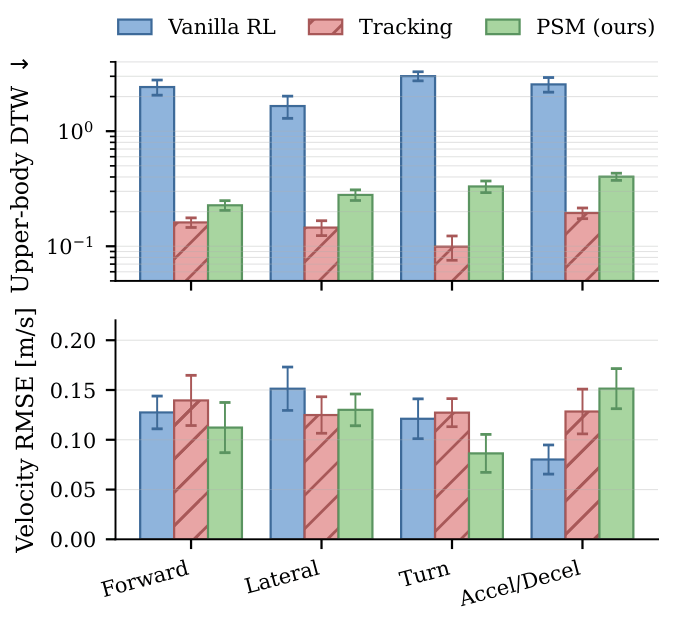}
    \caption{Naturalness vs.\ command scenario (simulation batch, no pushes). Top: upper-body DTW to the reference motion (log scale). Bottom: velocity RMSE to $\mathbf{u}(t)$ as a sanity check that all methods solve the same locomotion task.}
    \label{fig:naturalness-cmds}
  \end{figure}

  Each method is rolled out on a shared library of open-loop velocity schedules $\mathbf{u}(t) = (v_x, v_y, \omega_z)$ in the yaw frame.
  For naturalness trials, schedules are the root velocities of held-out walking clips from $\mathcal{D}$; additional scripted schedules (constant forward walking, lateral sinusoids, heading changes, accelerate--decelerate segments) are used for stress testing and robustness.
  Policies track the prescribed $\mathbf{u}(t)$ while we log joint and root states; the command is not resampled mid-episode, so observed differences reflect whole-body style and recovery rather than different command distributions.
  For style alignment we compare logged upper-body joints and gait scalars against the source reference clip that generated $\mathbf{u}(t)$ using dynamic time warping (DTW)~\cite{sakoe1978dynamic}, which tolerates mild phase shifts after turns or speed changes that would otherwise inflate a frame-wise RMSE:
  \begin{equation}
    \mathrm{DTW}(\mathbf{q}, \mathbf{r}) = \min_{\pi \in \mathcal{A}} \sum_{(i,j) \in \pi} d\!\left(\mathbf{q}_i, \mathbf{r}_j\right),
    \label{eq:dtw}
  \end{equation}
  where $\mathcal{A}$ denotes admissible warping paths and $d(\cdot,\cdot)$ is the squared joint error weighted as in~\eqref{eq:r-up} for the upper body.
  Scores are averaged over trajectories and random seeds (lower is better).
  Since all methods are scored against the same reference motions from $\mathcal{D}$, the DTW numbers are directly comparable: Tracking optimizes directly against that clip and serves as a tight lower bound on what is achievable under that supervision, while the gap between vanilla RL and PSM measures how far predictive matching moves a policy-only controller toward the human trajectory.

  The same $\mathbf{u}(t)$ trajectories are then repeated under a fixed schedule of impulsive pushes and velocity kicks.
  We report fall rate (\%), velocity-tracking error to $\mathbf{u}(t)$ in the post-disturbance window, and task recovery time $T_{\mathrm{vel}}$ (the time, in seconds, until $\|\mathbf{u}_t - \hat{\mathbf{v}}_t\| < \epsilon$).
  Task quantities use time-aligned errors.
  With protocol and metrics fixed, we first isolate naturalness in the absence of disturbances, then introduce pushes and quantify what each method gives up.

  \subsection{Naturalness}
  \label{sec:naturalness-results}

  We begin with nominal walking, where the policy is not driven into recovery poses.
  Fig.~\ref{fig:naturalness-cmds} reports DTW to the reference motion across the no-push trajectory batch in simulation, split by command scenario.
  Tracking attains the lowest upper-body and gait DTW because it directly optimizes against the source clip used for scoring---an expected lower bound for that supervision setting.
  Aggregated over the batch, PSM reduces upper-body DTW by roughly $8{\times}$ relative to vanilla RL ($0.31 \pm 0.03$ vs.\ $2.41 \pm 0.30$), and gait DTW by about $4{\times}$ ($0.21 \pm 0.04$ vs.\ $0.81 \pm 0.20$).
  PSM remains above Tracking because it follows predictor outputs recomputed from the policy's own lower-body history and command rather than the source clip itself---precisely the property that decouples style from a clip clock when recovery is needed.

    \begin{figure}[!t]
    \centering
    \includegraphics[width=\columnwidth]{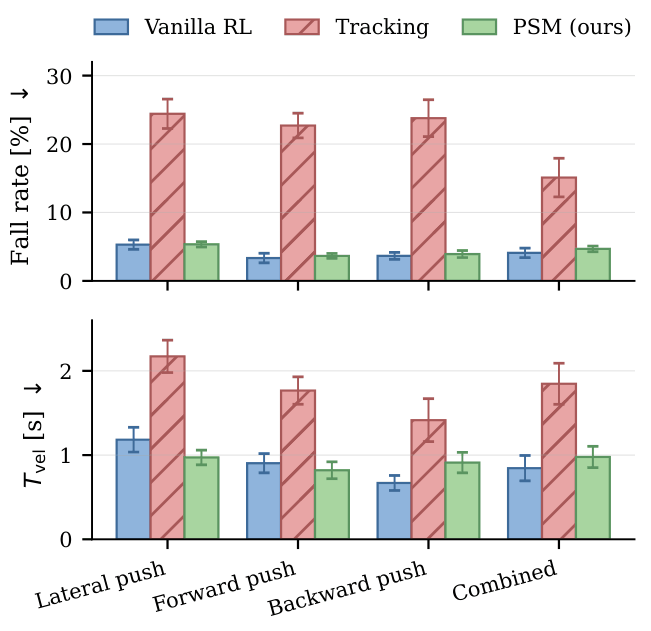}
    \caption{Robustness vs.\ disturbance scenario (simulation batch). Top: fall rate. Bottom: task recovery time $T_{\mathrm{vel}}$.}
    \label{fig:robustness-cmds}
  \end{figure}
  Tracking remains the lower-bound method when its reference phase is available, whereas PSM closes most of the gap from vanilla RL without requiring that reference at deployment.
  Velocity-tracking RMSE to $\mathbf{u}(t)$ is comparable across methods, confirming that the style differences are not a byproduct of solving different locomotion tasks.
  On hardware all three policies execute the same representative $\mathbf{u}(t)$ subset, and PSM produces the coordinated arm swing and stepping shown in Fig.~\ref{fig:psm-hardware}.

  \subsection{Robustness}
  \label{sec:robustness-results}

\begin{figure}[t]
    \centering
    \includegraphics[width=\columnwidth]{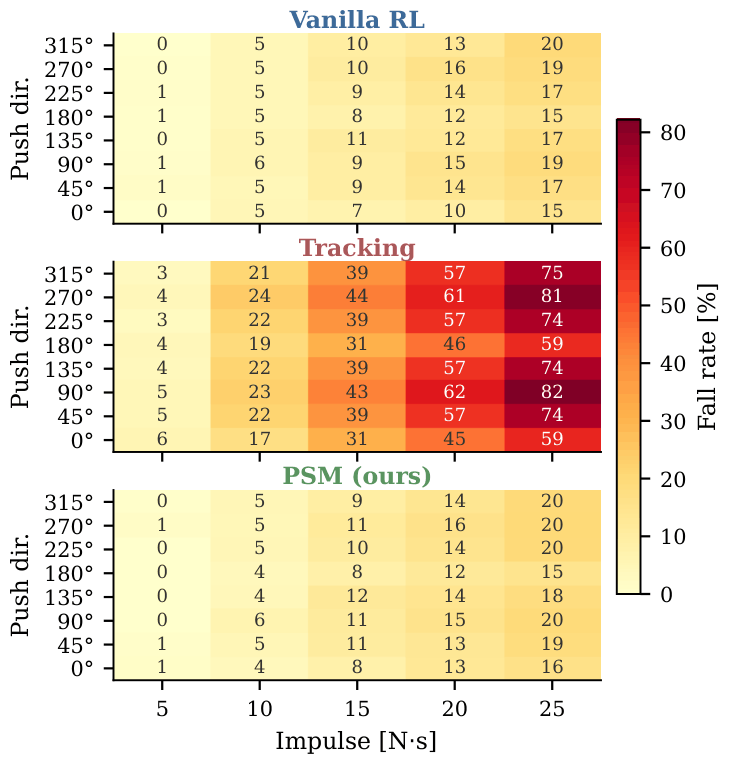}
    \caption{Push envelope (simulation): fall rate over push direction and impulse magnitude (shared color scale). Vanilla RL and PSM remain low across the envelope, while Tracking saturates at moderate impulses, particularly for lateral pushes.}
    \label{fig:disturbance}
  \end{figure}

  Naturalness alone is only half the picture: a controller that walks well only when nothing pushes back is of little practical use.
  We therefore repeat the same command trajectories under impulsive pushes and velocity kicks.
  Fig.~\ref{fig:robustness-cmds} breaks down fall rate and recovery time by disturbance type, and Fig.~\ref{fig:disturbance} summarizes the push envelope over direction and impulse magnitude.
  Aggregated across the disturbance suite, vanilla RL and PSM remain within one standard deviation of each other on both fall rate ($4.1\%$ vs.\ $4.4\%$) and task recovery $T_{\mathrm{vel}}$ ($\sim$0.9\,s), whereas Tracking incurs roughly $5{\times}$ the fall rate ($21.5\%$) and the longest $T_{\mathrm{vel}}$ ($1.80$\,s).
  This pattern is consistent with clip targets opposing the transient recovery poses that the task reward is simultaneously trying to elicit.

  On hardware, \href{https://www.youtube.com/watch?v=KmXUvEOFiy4}{PSM exhibits strong push robustness} while retaining natural, human-like arm motion.
  Taken together, Figures~\ref{fig:naturalness-cmds} and~\ref{fig:robustness-cmds} sketch a clear naturalness--robustness trade-off: PSM sits next to vanilla RL on the fall-rate axis while reducing upper-body DTW by close to an order of magnitude, whereas Tracking occupies the opposite corner---lowest DTW but highest fall rate---of the same plane.
  Beyond the headline numbers, a development-time ablation is worth noting: activating only the upper-body weight $w_{\mathrm{up}}$ already biased stepping rhythm through kinematic coupling between arms, torso, and pelvis; the reported policies nevertheless match both upper-body and gait descriptors, because explicit gait matching proved more stable across command and disturbance scenarios.

  Overall, the main gain of PSM is not only lower style error, but a cleaner decomposition of responsibilities during learning: locomotion rewards dominate balance and recovery transients, while predictor-based terms act as local style attractors for upper-body coordination and gait timing.
  This appears to explain the observed operating point: PSM moves markedly toward human-like motion under nominal commands while remaining close to vanilla RL under pushes, whereas clip-tracking reaches lower nominal DTW but pays a larger robustness cost when recovery requires off-reference poses.
  Practically, this keeps deployment simple: only $\pi_\theta$ runs on hardware, with no clip index, phase-estimation pipeline, or additional runtime conditioning channel.
  The disturbance breakdown is especially informative: Tracking degrades most under lateral impulses, where recovery poses diverge farthest from the reference clip.
  In that regime, state-conditioned predictor targets remain aligned with the task objective rather than pulling the policy back toward a nominal walking pose.

  %%%%%%%%%%%%%%%%%%%%%%%%%%%%%%%%%%%%%%%%%%%%%%%%%%%%%%%%%%%%%%%%%%%%%%%%%%%%%%%%
  \section{Conclusion and Future Work}
  \label{sec:conclusion}

  Taken together, the naturalness and robustness experiments support the central design choice of PSM: use human motion as a \emph{train-time}, state-conditioned style signal rather than a deployment-time reference.
  Because matching is consumed only during training, the deployed controller retains the proprioceptive interface and inference cost of a task-only RL baseline, while the lower-body-conditioned predictor supplies RL with a concrete notion of arm swing, stepping, and torso behavior that task rewards alone do not specify.
  In the controlled G1 comparison, PSM reduces upper-body DTW by roughly $8{\times}$ relative to vanilla RL while staying within one standard deviation on both fall rate and $T_{\mathrm{vel}}$.
  The clip-tracking baseline attains the lowest DTW against its own reference, yet falls about $5{\times}$ more often when those references oppose recovery---a behavior that follows from training-time choices (locomotion-first curriculum, pushes throughout training, state-conditioned targets) rather than any runtime stabilizer.

  Several limitations are worth flagging: predictor quality upper-bounds the achievable style, the sim-to-mocap leg-state shift is not corrected explicitly, and only the first predicted step ($k{=}0$) is used for matching even though $f_\phi$ produces a full horizon.
  Natural next steps include matched AMP~\cite{peng2021amp,escontrela2022adversarial} and universal-tracker~\cite{sleiman2026zest} comparisons under the same $\mathcal{D}$ and MDP, fine-tuning $f_\phi$ on RL rollouts to close the sim-to-mocap loop, multi-step anticipatory matching that exploits the full predicted horizon, terrain- or contact-conditioned predictors beyond flat-ground walking, and transferring the same recipe to other humanoid platforms whose kinematics differ from the G1 (e.g., taller bipeds or robots with different arm and torso topologies), where only $f_\phi$ and the matched joint subset need to be re-instantiated.

  %%%%%%%%%%%%%%%%%%%%%%%%%%%%%%%%%%%%%%%%%%%%%%%%%%%%%%%%%%%%%%%%%%%%%%%%%%%%%%%%

  \bibliographystyle{IEEEtran}
  \bibliography{ref}

  \end{document}